\PassOptionsToPackage{dvipsnames,table}{xcolor}
\documentclass[10pt,twocolumn,letterpaper]{article}

\usepackage[pagenumbers]{iccv} 
\usepackage{multirow}
\usepackage{graphicx}
\usepackage{booktabs}
\usepackage{graphicx}
\usepackage{amsmath}
\usepackage{amssymb}
\usepackage{booktabs}
\usepackage{multirow}
\usepackage{graphicx}
\usepackage{booktabs}
\usepackage{algorithm}
\usepackage{algorithmic}
\usepackage{xcolor}

\definecolor{iccvblue}{rgb}{0.21,0.49,0.74}
\usepackage[pagebackref=true,breaklinks=true,letterpaper=true,colorlinks,bookmarks=false]{hyperref}
\usepackage[table]{xcolor} 

\def\paperID{12655} 
\def\confName{ICCV}
\def\confYear{2025}

\begin{document}
\title{OmniCache: A Trajectory-Oriented Global Perspective on Training-Free Cache Reuse for Diffusion Transformer Models}
\author{\authorBlock}
\author{Huanpeng Chu\footnotemark[1]\\
Zhipu AI\\
{\tt\small chuhp@zju.edu.cn}
\and
Wei Wu\footnotemark[1]\\
Nanjing University\\
{\tt\small 714031909@qq.com}
\and
Guanyu Fen\footnotemark[2]\\
Zhipu AI\\
{\tt\small
guanyu.feng@zhipuai.cn}
\and
Yutao Zhang\\
Zhipu AI\\
{\tt\small z\_zhangyt@163.com
 }
}
\maketitle
\footnotetext[1]{Both authors contributed equally to this research.}
\footnotetext[2]{Corresponding author.}
\begin{abstract}

Diffusion models have emerged as a powerful paradigm for generative tasks such as image synthesis and video generation, with Transformer architectures further enhancing performance. However, the high computational cost of diffusion Transformers—stemming from a large number of sampling steps and complex per-step computations—presents significant challenges for real-time deployment. In this paper, we introduce OmniCache, a training-free acceleration method that exploits the global redundancy inherent in the denoising process.

Unlike existing methods that determine caching strategies based on inter-step similarities and tend to prioritize reusing later sampling steps, our approach originates from the sampling perspective of DIT models. We systematically analyze the model's sampling trajectories and strategically distribute cache reuse across the entire sampling process. This global perspective enables more effective utilization of cached computations throughout the diffusion trajectory, rather than concentrating reuse within limited segments of the sampling procedure.
In addition, during cache reuse, we dynamically estimate the corresponding noise and filter it out to reduce its impact on the sampling direction.

Extensive experiments demonstrate that our approach accelerates the sampling process while maintaining competitive generative quality, offering a promising and practical solution for efficient deployment of diffusion-based generative models. 
\end{abstract}

\begin{figure}[thbp]
    \centering
    \begin{subfigure}[t]{0.63\linewidth}
        \centering
        \includegraphics[width=\linewidth]{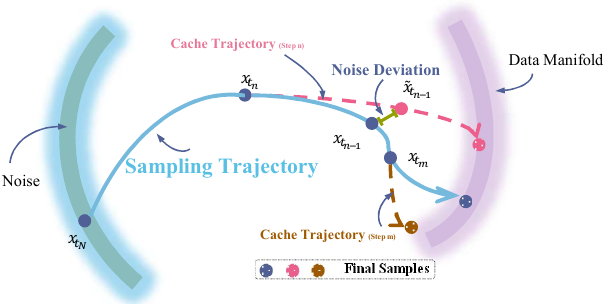}
        \caption{Sampling}
        \label{fig:sample}
    \end{subfigure}
    \hfill
    \begin{subfigure}[t]{0.35\linewidth}
        \centering
        \includegraphics[width=\linewidth]{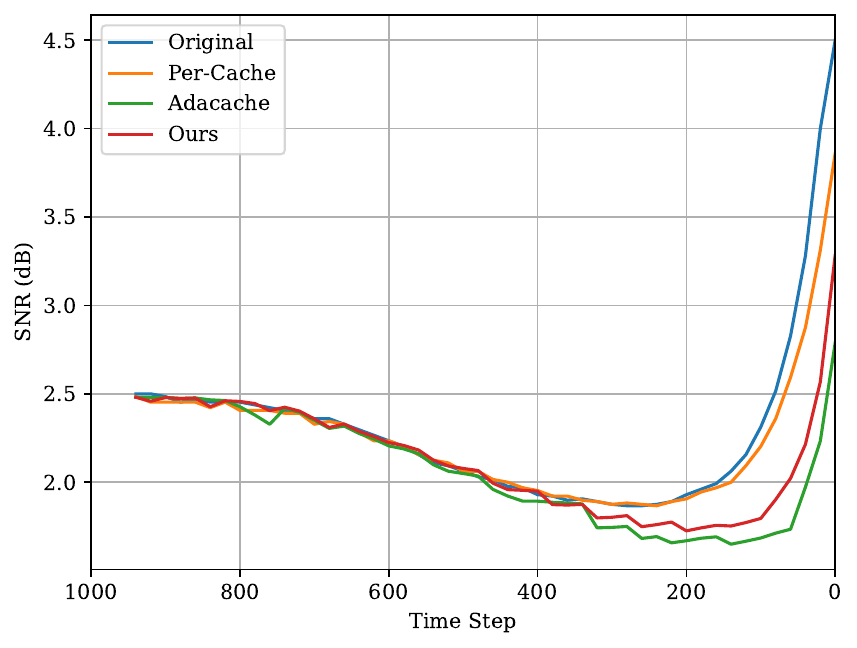}
        \caption{SNR Variation}
        \label{fig:snr}
    \end{subfigure}
    
    \caption{a). The geometric sampling model during diffusion illustrates that every initial sample (drawn from the noise distribution) begins from a large sphere and converges to the final sample along a regular sampling trajectory. We perform cache reuse at different time steps $x_{t_{n}}$ and $x_{t_{m}}$. We see early cache reuse allows subsequent sampling steps to filter introduced noise, correcting the trajectory. In contrast, late-stage cache reuse produces irrecoverable trajectory deviations.
    b). Real SNR curve during sampling phases under cache reuse conditions. Since model outputs show greater similarity in later sampling stages yet exhibit weaker denoising strength, cache reuse in later sampling steps leads to continuous SNR degradation.}
    \label{fig:intro}
\end{figure}

\section{Introduction}
In recent years, diffusion models \cite{song2020score,song2019generative,ho2020denoising} have achieved remarkable performance as 
powerful image generation models \cite{rombach2022ldm,dhariwal2021diffusion,wang2024ms}. Among the various backbone architectures for 
diffusion models, the Transformer \cite{vaswani2017attention} has demonstrated significant competitiveness, not 
only generating high-fidelity images \cite{peebles2023scalable,bao2023all}, but also excelling in video generation \cite{lu2023vdt,chen2023gentron,brooks2024video}, 
text-to-speech synthesis \cite{liu2023vitts,jing2023uditts,tan2024litefocus}, and 3D generation \cite{mo2024dit3d,cao20233d}. However, while diffusion 
Transformers benefit from the outstanding scalability of the Transformer architecture, 
they also bring substantial efficiency challenges, including high deployment costs and slow 
inference speeds. As sampling costs scale proportionally with both the number of time steps 
and the size of the model at each step, current approaches to improving sampling efficiency naturally 
split into two directions: reducing the number of sampling steps \cite{song2020ddim,ho2020denoising,li2024snapfusion,kim2023bksdm} or lowering the per-step 
inference cost \cite{fang2024structural,yang2023diffusionslim}.


Methods for reducing the number of time steps include distilling the trajectory to fewer steps \cite{salimans2022progressive,song2023consistency,luo2023latent} 
and discretizing the reverse-time stochastic differential equation (SDE) or the ordinary differential 
equation (ODE) of probability flows \cite{song2020ddim,zhang2022gddim,lu2022dpm}. On the other hand, the second approach primarily 
focuses on model compression \cite{kim2023bksdm,li2024snapfusion}, such as using model quantization to transform the 
denoising network into lower-bit representations \cite{he2023ptqd,shang2023postquant}.

A novel method for dynamic inference in diffusion models employs special caching mechanisms during 
the denoising process\cite{deepcache2023,wimbauer2023cacheme,learningtocache,adacache,tgate}. The starting point of these methods lies in the redundancy introduced 
by the diffusion model during sampling, which leads to high similarity between adjacent time steps. 
As a result, the computation for certain units can be reused in subsequent steps. However, as the 
noise is progressively filtered out, the more consistent inputs lead to higher similarity in denoising 
outputs. Therefore, existing caching strategies are more focused on later stages of the sampling process. 
However, output similarity is not an ideal metric for evaluating cache reuse, especially for current video generation models based on DIT. 

Fig.~\ref{fig:intro} provides a visual illustration and experimental validation of our approach. As shown in Fig.~\ref{fig:sample}, when cache reuse is implemented during early sampling stages, subsequent sampling steps naturally filter out the cache-induced noise—a fundamental characteristic of diffusion models. This allows the overall sampling trajectory to be gradually corrected toward the proper direction. In contrast, concentrating cache reuse in later steps leaves the model with insufficient capacity to remediate the introduced noise, resulting in irrecoverable trajectory deviations. Fig.~\ref{fig:snr} further substantiates our hypothesis, given that model outputs exhibit higher similarity during later sampling stages, which corresponds to diminished denoising strength. Consequently, noise introduced through cache reuse during these later phases cannot be effectively eliminated, leading to a progressive deterioration in the signal-to-noise ratio. This empirical evidence strongly supports our approach of distributed cache reuse throughout the entire sampling process rather than focusing exclusively on later steps.

To address these issues, we propose OmniCache,  a trajectory-oriented global perspective on training-free cache reuse for Diffusion Transformer Models.
First, we simplify and estimate the sampling trajectories of various diffusion models, revealing a pattern that is independent of the input sample. Leveraging the inherent regularity of these trajectory shapes, we introduce a curvature-based approach to determine the optimal nodes for cache reuse. Additionally, we estimate and correct the noise induced by cache reuse. Considering that diffusion models primarily rely on low-frequency signals during the early stages of sampling to establish the overall structure and basic semantics, and incorporate high-frequency signals in the later stages to capture finer details, 
we apply either high-pass or low-pass filtering to the estimated noise, thereby preventing unnecessary alterations  to the hidden states across different sampling stages.
Our method achieves a 2–2.5$\times$ speedup on models such as Opensora and Latte with virtually no loss in performance metrics. Moreover, on more challenging or less redundant distilled models (e.g., CogVideoX-5b-I2V-distill~\cite{yang2024cogvideox}, 16steps), our approach yields a 1.45$\times$ speedup without performance degradation, whereas existing methods tend to lead to model collapse.

The contributions of our paper are as follows:
\begin{itemize}
    \item We analyzed the influence of cache reuse on the sampling process of DIT models and pioneered a novel cache reuse methodology based on model sampling trajectories. To the best of our knowledge, this work represents the first systematic investigation that optimizes cache reuse strategies by explicitly considering the inherent characteristics of the diffusion sampling path.
    \item We discovered that noise introduced by cache reuse in the current step exhibits remarkably high correlation with the previous step. By leveraging this characteristic, we further estimate and filter out cache-induced nosie. In addition, we incorporate both high-pass and low-pass filtering on estimated noise artifacts to mitigate the impact on irrelevant signals.
    \item Our experimental results demonstrate that our approach delivers a 2–2.5$\times$ speedup on models such as Opensora and Latte without any performance degradation. Furthermore, we are able to further exploit the potential of cache reuse, achieving a 1.45$\times$ speedup on distilled models with extremely low redundancy.
\end{itemize}




\section{Related Work}

\subsection{Efficient diffusion models}
Diffusion models have broken the long-time dominance of generative adversarial networks (GANs~\cite{goodfellow2020generative})  in the challenging task of image synthesis. However, their extremely slow generation speed and large memory footprint constrain their widespread use for downstream tasks.
\par
A great deal of recent work has focused on speeding up the sampling process while also improving the quality of the resulting samples. 
These efficient sampling methods can be classified into two main categories~\cite{ulhaq2022efficient,wang2022diffusion}:  
Efficient Process Strategies (EPS), which suggest ways to improve the efficiency of diffusion models 
or speed up the sampling process, and Efficient Design Strategies (EDS), 
which recommend modifications to the design of baseline diffusion models. 
Konwledge distillation~\cite{luhman2021knowledge,hoogeboom2021autoregressive}, 
analytical trajectory estimation~\cite{bao2022estimating}, 
early-stop strategy~\cite{lyu2022accelerating}, 
and differential equation (DE)~\cite{song2020score} 
are designed as EPS to decrease the discretization step size.
For EDS, SnapFusion~\cite{li2024snapfusion} obtains an efficient UNet structure by network pruning, 
which accelerates the inference process of the diffusion model.
The model class Latent Diffusion Models (LDMs)~\cite{rombach2022ldm} 
provide efficient image generation from the latent space with a single network pass. 
Model acceleration techniques—such as model quantization~\cite{chu2024qncd} and sparse attention~\cite{xi2025sparse}—have also been incorporated into diffusion models.

\par
However, the cache reuse method lies somewhere in between these two approaches. It leverages the similarity between consecutive steps in the diffusion process by storing intermediate results from the denoising network modules in cache to reduce computation.

\subsection{Cache Mechanisms}

In large language models and multimodal models, KV caching is widely used \cite{kvcache2020,kvcache2021}. This method caches the key and value components from the attention mechanism so that they can be reused in subsequent attention computations, thereby accelerating inference. Additionally, prefix caching stores computed results for a fixed prefix, which can then be directly reused during generation without redundant calculations.

There are also several cache-based approaches in diffusion model generation. For example, Deepcache \cite{deepcache2023} expedites computation across adjacent steps in UNet networks by caching the output feature maps of upsampling blocks. $\Delta$-DIT~\cite{deltadit}  stores the residuals between several blocks in the diffusion model to accelerate computation. AdaCache ~\cite{adacache} dynamically adjusts the intensity of cache reuse for each video being generated, while TOCA~\cite{toca} applies cache reuse along the token dimension in DIT. Furthermore, TGATE \cite{tgate2024} accelerates the cross-attention module in the later stages of sampling by caching output feature maps.

However, most existing methods are based on the similarity between adjacent steps and lack a comprehensive study on the impact of cache reuse on the sampling process. Our approach analyzes the effects of cache reuse based on the sampling trajectory and correct the noise induced by cache reuse in sampling stage.





\section{Method}
\begin{figure*}
    \centering
    \includegraphics[width=0.86\linewidth]{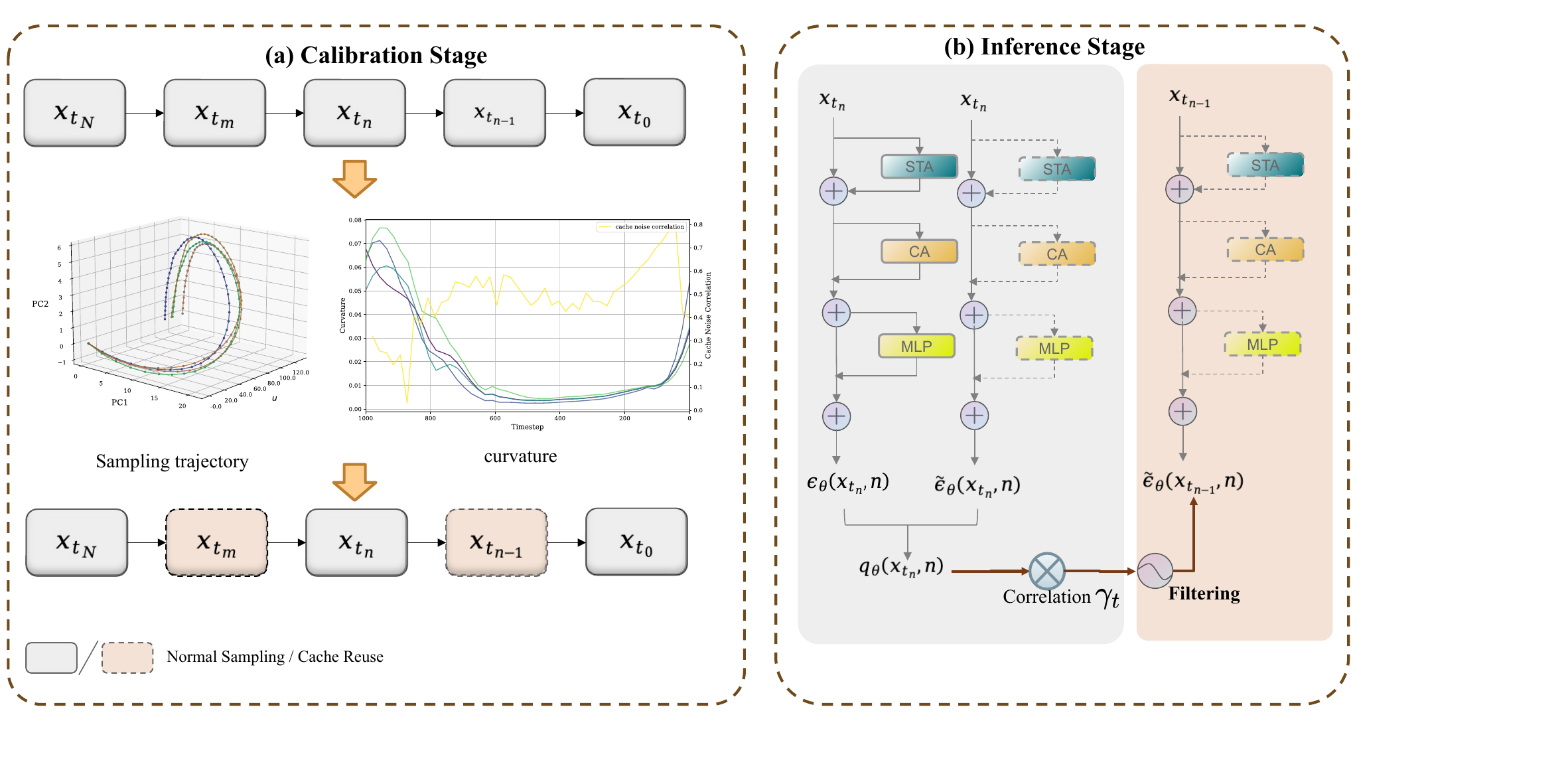}
    \caption{The diagram of our OmniCache. In the calibration stage, we store the states at different time steps \(x_{t_{n}}\) to obtain the corresponding sampling trajectory, while simultaneously computing the curvature and the correlation of cache-induced noise. Based on the curvature, we determine the appropriate time steps for cache reuse. In the actual sampling stage, we estimate the noise induced by cache reuse in real time, and correct the denoising model's output based on the noise correlation and high-pass/low-pass filtering.}
    \label{fig:pipeline}
\end{figure*}

In Section 3.1, we introduce the principles of the diffusion model's sampling process as well as the logic behind cache reuse. In Section 3.2, we model the sampling trajectory, analyze the impact of cache reuse on the trajectory, and propose a curvature-based cache reuse strategy. In Section 3.3, we decouple the noise induced by cache reuse and present a noise correction module. Finally, in Section 3.4, we provide a detailed description of our OmniCache method, covering both calibration stage and  sampling stage.

\subsection{Preliminaries}
Diffusion models are a family of probabilistic generative models that progressively destruct real data by injecting noise, then learn to reverse this process for generation, represented notably by  denosing diffusion probabilistic models (DDPMs~\cite{ho2020denoising}).  DDPM is composed of two chains: a forward chain that perturbs data to noise, and a reverse chain that converts noise back to data. The former is usually designed by hand and its goal is to convert any data distribution into a simple prior distribution (e.g., a standard Gaussian distribution).

\par
In the denoising process, with a Gaussian noise $x_T$, the diffusion model can generate samples by iterative sampling $x_{t-1}$ from $p_{\theta}(x_{t-1}|x_{t})$ until obtaining $x_0$,
where the Gaussian distribution  $p_{\theta}(x_{t-1}|x_{t})$ is a simulation of the unavailable real distribution  $q(x_{t-1}|x_{t})$. 
The mean value $\mu_{\theta}(x_{t-1}|x_{t})$ of $p_{\theta}(x_{t-1}|x_{t})$ is calculated from the noise prediction network $\epsilon_{\theta}$ (usually the UNet model or DiT model).
Therefore, the sampling process of $x_{t-1}$ is shown as follows:
\begin{equation}
\hspace{-0.4cm}
    \quad x_{t-1}=\frac{1}{\sqrt{\alpha_{t}}} \bigg (x_{t}-\frac{\beta_t}{\sqrt{1-\overline{\alpha_t}}} {\epsilon_{\theta}(x_{t},t) \bigg)}+\sigma_{t}z, \quad z \in N(0,I) \\
\end{equation}
The diffusion model continuously evokes the noise prediction network to acquire the noise $\epsilon_{\theta}(x_{t},t)$ and filter it out.
The complexity of the noise prediction network $\epsilon_{\theta}$ and the huge hidden states especially for video-generation 
make the sampling of diffusion models expensive.

The noise prediction network $\epsilon_{\theta}$, taking the current mainstream video generation model architecture DIT (Latent Diffusion Transformers) as an example, is composed of multiple layers stacked together (for instance, Latte-1 consists of 48 layers with identical structures).  
Each layer comprises an attention module (Atten) and a linear layer (MLP) arranged in a residual format. The essence of cache reuse is to store the output of the previous step for use in the current step (using the attention module as an example):
\begin{equation}
    \widetilde{f}_{t}^{l} = \begin{cases}
        {f}_{t}^{l} + Atten({f}_{t}^{l})   & Original Forward \\
        {f}_{t}^{l} + AttenCache &  Cache Reuse
\end{cases}
\end{equation}

For the input feature \({f}_{t}^{l}\) at the current step, the cache reuse strategy employs the previously stored \(AttenCache\) as the output of the attention module, thereby saving the corresponding computation. Typically, \(AttenCache\) is obtained by storing the output of the attention module from the previous step \(t-1\) or even earlier steps.
The discrepancy between \(AttenCache\) and the real output \(Atten({f}_{t}^{l})\) accumulates through subsequent layers, ultimately introducing unavoidable noise in the model's final output.


\begin{equation}
\begin{split}
   \hspace{-0.4cm}
        \quad \widetilde{x}_{t-1}&\!=\frac{1}{\sqrt{\alpha_{t}}} \bigg (x_{t}\!-\frac{\beta_{t}}{\sqrt{1\!-\overline{\alpha_t}}} {\widetilde{\epsilon}_{\theta}(x_{t},t) \bigg)}+\sigma_{t}z, \! \quad z \! \in N(0,I) \\
&\!=\frac{1}{\sqrt{\alpha_{t}}} \bigg (x_{t}\!-\frac{\beta_{t}}{\sqrt{1\!-\overline{\alpha_t}}} \big({{\epsilon}_{\theta}(x_{t},t) +q_\theta(x_{t},t)\big) \bigg)}+\sigma_{t}z. 
\label{eq:cache_noise}
\end{split}
\end{equation}

From the perspective of complete sampling process,
these cache noises accumulate to form  cache noise $q_\theta(\widetilde{x}_{t},t)$ which further accumulates in the current output $\widetilde{x}_{t-1}$ , thus affecting subsequent sampling processes.




\subsection{Sampling Trajectory of Diffusion Models}  
Diffusion models generate samples that conform to the true data distribution by injecting noise into the data and subsequently removing it via a reverse process. This process is typically implemented by solving the reverse diffusion equation (e.g., PF-ODE). During the sampling phase, the process of generating a sample from the noise space can be regarded as a trajectory between the noise distribution and the data distribution, i.e., the sampling trajectory. Closely related to this is the denoising trajectory, which is produced at each denoising step and implicitly influences the curvature and shape of the sampling trajectory. Specifically, the denoising trajectory determines the rotation direction at each sampling point, and this rotational information reflects the curvature of the sampling trajectory.

Existing methods~\cite{chen2024trajectory} provide an approach to approximate the sampling trajectory by confidently using a 3-D subspace.
First, we compute the unit vector between the starting and ending points:
\begin{equation}
\vspace{-0.1cm}
    u = \frac{x_{t_N} - x_{t_0}}{\|x_{t_N} - x_{t_0}\|} 
    \vspace{-0.1cm}
\end{equation}

Next, we project the trajectory onto the orthogonal complement of \(u\). We perform PCA on this projection and extract two principal directions, \(w_1\) and \(w_2\). Each sampling point is then approximated in the 3-D subspace spanned by \(u\), \(w_1\), and \(w_2\):
\begin{equation}
    {x}_{t_i}^{\mathrm{proj}} = ({x}_{t_i}\cdot u)u + ({x}_{t_i}\cdot w_1)w_1 + ({x}_{t_i}\cdot w_2)w_2.
\end{equation}

These two additional principal components allow us to more accurately approximate the actual trajectory, with deviations from the true trajectory being almost indistinguishable. This enables a better understanding of the geometric structure of high-dimensional sampling trajectories.

\subsubsection{Sampling Trajectories Exhibit a Boomerang Shape}  
1. As shown in Fig.~\ref{fig:cogvideo_sample}, the scale along the \(x_{t_{N}} - x_{t_0}\) axis is significantly larger than that of the other two principal components. Consequently, the trajectory is very close to the straight line connecting its endpoints.  
2. Trajectories from different initial starting points all exhibit a consistent boomerang shape, demonstrating strong regularity that is independent of the specific generated content and inherent solely to the diffusion model itself.  
3. The spacing between adjacent steps along the sampling trajectory gradually decreases. The intervals between trajectory points essentially reflect the differences between adjacent steps. Therefore, similarity-based cache reuse methods tend to occur more frequently in the later stages of sampling.

\subsubsection{Impact of Cache Reuse on  Sampling Trajectory}  

\begin{figure}
    \centering
    \includegraphics[width=0.95\linewidth]{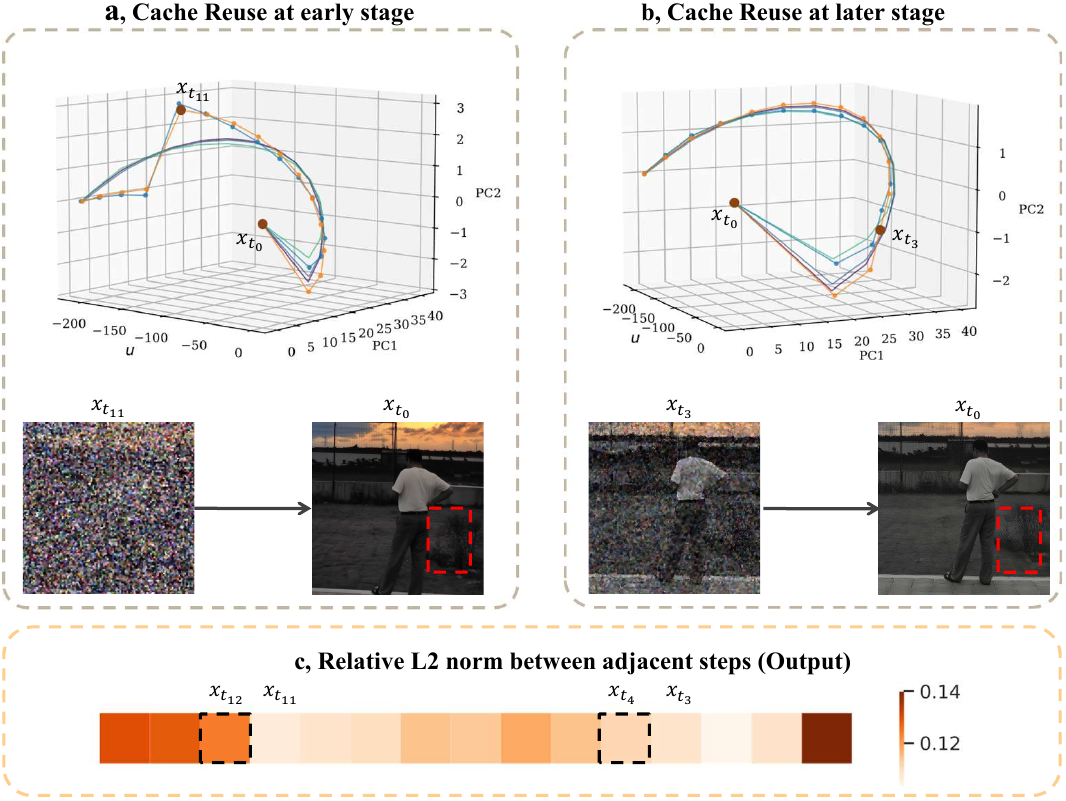}
    \caption{We visualized the sampling trajectories of the distilled version of CogVideoX-5b-I2V-distill~\cite{yang2024cogvideox} (a total of 16 steps). The unmarked trajectory in a and b represents the normal sampling process. Additionally, we applied cache reuse at one step in the early stage (left panel) and one step in the later stage (right panel), and visualized the corresponding intermediate outputs along with the final output results. In c, we visualized the relative L2 error between the output noise at each step and that of the previous step.}
    \label{fig:cogvideo_sample}
    \vspace{-0.3cm}
\end{figure}

In order to more intuitively demonstrate the impact of cache reuse on the sampling trajectory, we conducted an analysis on the CogVideoX-5b-I2V-distill model. Since CogVideoX-5b-I2V-distill utilizes only 16 sampling steps, the effects of cache reuse become even more pronounced. In Fig.~\ref{fig:cogvideo_sample}a, we applied cache reuse during an early stage of sampling ($x_{t_{12}}\rightarrow x_{t_{11}}$). At this point, it is evident that cache reuse itself introduces additional noise, which misleads the sampling direction of the diffusion model. However, the diffusion model has an inherent self-correction capability. During the early sampling phase, when cache reuse is applied, subsequent sampling steps treat the noise introduced as part of the denoising process, thereby pulling the sampling direction back on track. As shown in Fig.~\ref{fig:cogvideo_sample}a, the hidden states $x_{t_{11}}$ are dominated by noise, so the additional noise introduced can be effectively ignored. In contrast, during the later sampling stage ($x_{t_{4}}\rightarrow x_{t_{3}}$), the noise induced by cache reuse is much harder to disregard, and there are not enough subsequent sampling steps to correct it, making the change in sampling direction irreversible.

Furthermore, we visualized the time steps at which cache reuse was applied along with the final generated outputs. It can be observed that when cache reuse is applied at $x_{t_{4}}$, the generated results exhibit \textbf{ripple-like noise regions} (highlighted by red boxes), leading to poorer quality compared to applying cache reuse at  $x_{t_{12}}$. This phenomenon is consistent with the SNR plot in Fig.~\ref{fig:snr}: introducing additional noise into the noise-dominated hidden states has little effect during the early sampling stages, whereas the impact is pronounced in the later stages.


However, as shown in Fig.~\ref{fig:cogvideo_sample}c, the relative L2 norm of the output noise at \(x_{t_4}\) is lower than that at \(x_{t_{12}}\). Therefore, similarity-based cache reuse strategies tend to select \(x_{t_4}\) for cache reuse (as in AdaCache). Consequently, the locally based similarity metric of output features diverges from the truly optimal selection strategy. We, instead, explore a more comprehensive cache reuse strategy based on the sampling trajectory.



\subsubsection{Trajectory Curvature-Based Cache Reuse}

Our analysis reveals that sampling steps with higher similarity predominantly emerge in  later stages of the sampling process, precisely when cache reuse introduces irreversible perturbations to the sampling trajectory. Based on this insight, we propose a novel caching strategy grounded in the global properties of sampling trajectories.
We identify that trajectory curvature serves as an effective indicator for optimal cache deployment: regions of minimal curvature signify stable directional progression, making these ideal candidates for cache reuse without compromising generation quality. Conversely, segments exhibiting pronounced curvature represent critical directional transitions where standard sampling processes should be preserved.
Notably, our comprehensive experiments demonstrate remarkable structural consistency across diverse sampling trajectories, with nearly identical geometric patterns emerging regardless of generation contents. This consistency enables our curvature-based caching methodology to generalize effectively across various video generation tasks.


\subsection{Cache-induced Noise Correction}  
The noise induced by a single step of cache reuse in Eqn.~\ref{eq:cache_noise} can be computed as follows:
\begin{equation}
\vspace{-0.1cm}
    \begin{split}
    q_\theta(x_t,t) &= \widetilde{\epsilon}_{\theta}(x_{t},t) -\epsilon_{\theta}(x_{t},t) 
    \end{split} 
\vspace{-0.1cm}
\end{equation}
When we perform a Taylor expansion on cache-induced noise $q_\theta(x_t,t) $, we obtain the following formula:
\begin{equation}
    \begin{split}
    q_\theta(x_{t-1},t-1) &\approx q_\theta(x_{t},t) - \frac{d\,q_\theta(x_t,t)}{dt}   + \mathcal{O}(\Delta t^2)  \\
    &\approx \gamma_{t-1}\ q_\theta(x_t,t),
    \end{split} 
    \label{eqn:noise_corr}
\end{equation}
where $\gamma_{t-1}$ is the correlation of noise. 

This indicates that when we perform normal forward sampling in the preceding step, we can simultaneously perform cache reuse to obtain the corresponding cache reuse noise $q_\theta(x_t,t)$ for estimating $q_\theta(x_{t-1},t-1)$ .
In order to estimate the noise induced by cache reuse more accurately, we need to minimize the second term in Eqn.~\ref{eqn:noise_corr}. When we further decouple the second term, we obtain:
\begin{equation}
    \begin{split}
    -\frac{d\,q_\theta(x_t,t)}{dt} & = \frac{d\,\epsilon_{\theta}(x_{t},t)}{dt}-\frac{d\,\widetilde{\epsilon}_{\theta}(x_{t},t)}{dt}   \\ 
    & = \frac{\sqrt{\alpha_t-\alpha_t\overline{\alpha_t}}}{\beta_t} \bigg (\frac{d\,\widetilde{x}_{t-1}}{dt} -\frac{d\,x_{t-1}}{dt} \bigg) \\
    & \approx \frac{\sqrt{\alpha_t-\alpha_t\overline{\alpha_t}}}{\beta_t} \bigg (
    \frac{d\,{x}_{t}}{dt} - \frac{d\,x_{t-1}}{dt}  \bigg ) \\
    & \approx \frac{\sqrt{\alpha_t-\alpha_t\overline{\alpha_t}}}{\beta_t} \frac{d^2 x_t}{dt^2}
    \end{split} 
\end{equation}

The first derivative of $q_\theta(x_t,t)$ is correlated to the second derivative of the sampling trajectory. That is, when we select a time step with minimal curvature for cache reuse, the first derivative of the cache noise is minimized, and at this point the actual cache noise $q_\theta(x_{t-1},t-1)$  is most strongly correlated with the cache noise $q_\theta(x_{t},t)$  from the previous step. 
As shown in Fig.~\ref{fig:pipeline}, we visualize the variation of curvature and noise correlation $\gamma_t$ over time steps in the Latte model. It can be observed that curvature and $\gamma_t$ are negatively correlated, which is consistent with our previous predictions. When the curvature is low, the sampling trajectory tends to continue along the previous direction, resulting in a stronger correlation in the noise introduced by cache reuse.



Therefore, we can use the estimated cache noise  $\gamma_{t-1}\ q_\theta(x_t,t)$ to correct the current step.
Besides, our trajectory curvature-based cache reuse and the correction of cache-induced noise are inherently coupled and mutually reinforcing.

Furthermore, for diffusion models with high redundancy (for example, 50-step diffusion models such as Latte), there can be consecutive time steps where cache reuse is applied. In this case, the estimated noise can be approximated as the product of multiple correlation coefficients:
\begin{equation}
    \begin{split}
    q_\theta(x_{t-1},t-1) 
    &\approx \gamma_{t-1}\ q_\theta(x_t,t) \\
    & \approx \gamma_{t-1} * \gamma_{t} \ q_\theta(x_{t+1},t+1)
    \end{split} 
\end{equation}
Therefore, to ensure the effectiveness of the noise correction module, we impose an additional constraint that prohibits performing cache reuse for three consecutive steps.

At the same time, considering that diffusion models process different signals at different sampling stages—focusing on global information and low-frequency signals in the early stages, and on fine details and high-frequency signals in the later stages. We apply low-pass filtering to the estimated noise during the early sampling stages to prevent it from introducing additional noise into the high-frequency components. Conversely, in the later stages, we apply high-pass filtering to prevent low-frequency noise from degrading the underlying information.

\subsection{Summary of Methods}  
As shown in the Fig.~\ref{fig:pipeline}, our method consists of the following steps:
\begin{itemize}
    \item During the \textbf{calibration stage}, we obtain the input state $x_{t}$ of  diffusion model at each step,  output $\epsilon_{\theta}(x_{t},t)$ by the denoising model.
    Then, we reduce the sampling trajectory to three dimensions and calculate the curvature at each step. Besides, we must ensure that no three consecutive time steps are selected for cache reuse, so as to guarantee the effectiveness of subsequent noise correction.
    \item During the \textbf{actual sampling stage}, when a predetermined cache reuse time step is reached, we replace the original computed results with the cached attention outputs and MLP outputs, thereby saving computational cost.
    Based on the noise correlation scale $\gamma_t$, we estimate the cache-induced noise, then apply either high-pass or low-pass filtering to correct the output of the denoising model.
\end{itemize}

\begin{table*}[htbp]
\centering
\resizebox{0.9\textwidth}{!}{%
\begin{tabular}{@{}l|c|c|c|c|c|c@{}}
\toprule
\textbf{Method} & \multicolumn{1}{l|}{\textbf{VBench (\%) $\uparrow$}} & \multicolumn{1}{l|}{\textbf{PSNR $\uparrow$}} & \multicolumn{1}{l|}{\textbf{LPIPS $\downarrow$}} & \multicolumn{1}{l|}{\textbf{SSIM $\uparrow$}} & \multicolumn{1}{l|}{\textbf{FLOPs (T)}} & \textbf{Speedup} \\ \midrule
Open-Sora            & 79.22  & --    & --     & --     & 3230.24 & 1.00$\times$ \\ \midrule
+ $\Delta$-DiT         & 78.21  & 11.91 & 0.5692 & 0.4811 & 3166.47 & --          \\
+ T-GATE            & 77.61  & 15.50 & 0.3495 & 0.6760 & 2818.40 & 1.10$\times$ \\
+ PAB-fast             & 76.95  & 23.58 & 0.1743 & 0.8220 & 2558.25 & 1.34$\times$ \\
+ PAB-slow            & 78.51  & 27.04 & 0.0925 & 0.8847 & 2657.70 & 1.20$\times$ \\
+ ToCa(R = 85\%) &78.34 &-- &-- &-- &1394.03 & 2.36$\times$ \\
+ TeaCache-fast &78.48 &19.10 &0.2511 &0.8415 &1640.00 &2.25$\times$ \\
\rowcolor{gray!25}
+ OmniCache-slow  & 78.83 &22.37 &0.1553 &0.8180 &1615.12 &2.00$\times$ \\
\rowcolor{gray!25}
+ OmniCache-fast  & 78.50 &21.27 &0.1841 &0.7930 &1292.10 &2.50$\times$ \\
 \midrule
Latte                     & 77.40  & --    & --     & --     & 3439.47 & 1.00$\times$ \\ \midrule
+ $\Delta$-DiT       & 52.00  & 8.65  & 0.8513 & 0.1078 & 3437.33 & --          \\
+ T-GATE            & 75.42  & 19.55 & 0.2612 & 0.6927 & 3059.02 & 1.11$\times$ \\
+ PAB-fast              & 73.13  & 17.16 & 0.3903 & 0.6421 & 2576.77 & 1.33$\times$ \\
+ PAB-slow              & 76.32  & 19.71 & 0.2699 & 0.7014 & 2767.22 & 1.24$\times$ \\
+ AdaCache-fast                           & 76.26  & 17.70 & 0.3522 & 0.6659 & 1010.33 & 2.74$\times$ \\
+ AdaCache-fast (w/ MoReg)                  & 76.47  & 18.16 & 0.3222 & 0.6832 & 1187.31 & 2.46$\times$ \\
+ AdaCache-slow                           & 77.07  & 22.78 & 0.1737 & 0.8030 & 2023.65 & 1.59$\times$ \\ 
+ TeaCache-fast &76.69 &18.62 &0.3133 &0.6678 &1120.00 &3.28 $\times$ \\
\rowcolor{gray!25}
+ OmniCache-slow  & 77.24 &22.48 &0.1955 &0.7903 &1719.74 &2.00$\times$ \\
\rowcolor{gray!25}
+ OmniCache-fast  & 77.09 &21.06 &0.2463 &0.7575 &1375.79 &2.50$\times$ \\
\bottomrule
\end{tabular}%
}
\caption{Quantitative evaluation of text-to-video generation quality. Here, we compare our method, OmniCache, with several training-free, cache-based DiT acceleration approaches on multiple video baselines. It can be observed that our OmniCache-fast scheme strikes a favorable balance between the acceleration ratio and actual performance metrics, incurring only minimal performance degradation with a 2.5× speedup. Corresponding visualizations can be found in \color{red}{Appendix}.}
\label{tab:t2v}
\end{table*}

\section{Experiment}
\subsection{Implementation details}
\noindent
\textbf{Model Configurations} We conducted experiments across four widely-used DIT models for various generation tasks. In the image generation domain, DIT-XL~\cite{bao2023all} was employed for class-conditional image generation. For video generation scenarios, we implemented OpenSora~\cite{zheng2024opensora} (480p -2s at 30steps)  and Latte~\cite{latte}  (512$\times$512 -2s at 50steps) for text-to-video generation and CogVideoX-5b-I2V-distill~\cite{yang2024cogvideox} for image-to-video conversion. All experiments were performed on NVIDIA 4090 GPU hardware.

\noindent
\textbf{Evaluation Metrics} 
For class-conditional image generation, we uniformly sampled 1000 categories from ImageNet and generated 50000 samples at a resolution of 256 $\times$ 256 pixels. We then evaluated their FID \cite{heusel2017gans} and Precision and Recall metrics \cite{kynkaanniemi2019improved}. For text-to-video generation, we test our approach based on the VBench framework \cite{huang2024vbench}, generating 5 videos for each of the 950 benchmark prompts under different random seeds, resulting in a total of 4,750 videos. The generated videos are comprehensively evaluated across 16 aspects proposed in VBench. Besides, we also report reference-based PSNR, SSIM \cite{wang2004image} and LPIPS \cite{zhang2018perceptual} as quality metrics. For the image-to-video generation task, we randomly generated 100 videos and evaluated them using metrics such as Q-Align \cite{li2023qalign}. In addition, we conducted human evaluations to analyze and compare the videos across two major dimensions: consistency and video quality.

\noindent
\textbf{Comparison Targets} For benchmarking purposes, given that our proposed method is a training-free inference acceleration strategy, we selected several comparable approaches as baselines. The comparison methods include FORA~\cite{selvaraju2024fora}, $\Delta$-DiT~\cite{deltadit}, T-GATE~\cite{tgate}, PAB~\cite{pab}, AdaCache~\cite{adacache}, TeaCache~\cite{liu2024timestep} and ToCA~\cite{toca}. These methods represent the current state-of-the-art techniques for accelerating diffusion model inference without requiring additional training.

\begin{table}[tbhp]
\centering
\resizebox{0.4\textwidth}{!}{%
\begin{tabular}{lccc}
\toprule
\textbf{Method} & \textbf{Aesthetic Quality $\uparrow$} & \textbf{Q-Align$\uparrow$} & \textbf{Speedup} \\ \midrule
CogVideoX-5b-I2V-distill      & 0.59  & 0.79  & 1.00$\times$    \\
+AdaCache$^{*}$ & 0.57  & 0.77  & 1.33$\times$    \\
\rowcolor{gray!25}
+OmniCache     & 0.621 & 0.792 & 1.45$\times$    \\ \bottomrule
\end{tabular}%
}
\caption{Quantitative evaluation of image-to-video generation.}
\label{tab:cogvideo}
\end{table}

\begin{table*}[ht]
\centering
\resizebox{0.8\textwidth}{!}{%
\begin{tabular}{@{}lcccccc@{}}
\toprule
\multicolumn{1}{l|}{\textbf{Method}} & \multicolumn{1}{l}{\textbf{FLOPs(T)}} & \multicolumn{1}{l|}{\textbf{Speed}} & \multicolumn{1}{l}{\textbf{sFID$\downarrow$}} & \multicolumn{1}{l}{\textbf{FID$\downarrow$}} & \multicolumn{1}{l}{\textbf{Precision$\uparrow$}} & \multicolumn{1}{l}{\textbf{Recall$\uparrow$}} \\ \midrule
\multicolumn{1}{l|}{\textbf{DiT-XL/2-G (cfg = 1.50)}} & 118.68 & \multicolumn{1}{c|}{1.00×} & 4.98 & 2.31 & 0.82 & 0.58 \\ \midrule
\multicolumn{1}{l|}{33\% steps} & 39.40  & \multicolumn{1}{c|}{3.01×} & 6.31  & 2.76 & 0.81 & 0.57 \\
\multicolumn{1}{l|}{37\% steps} & 44.15  & \multicolumn{1}{c|}{2.69×} & 6.04  & 2.64 & 0.81 & 0.58 \\
\multicolumn{1}{l|}{+FORA (N = 3)} & 39.95  & \multicolumn{1}{c|}{2.97×} & 6.21  & 2.80 & 0.80 & 0.59 \\
\multicolumn{1}{l|}{+FORA (N = 2.8)} & 43.36  & \multicolumn{1}{c|}{2.74×} & 6.13  & 2.80 & 0.80 & 0.59 \\
\multicolumn{1}{l|}{+ToCa (N = 4, R = 93\%)} & 43.22  & \multicolumn{1}{c|}{2.75×} & 5.81  & 2.68 & 0.80 & 0.57 \\ \midrule
\rowcolor{gray!25}
\multicolumn{1}{l|}{+OmniCache-fast} & 47.47  & \multicolumn{1}{c|}{2.50×} & 5.86  & 2.69 & 0.81 & 0.56 
  \\ \bottomrule
\end{tabular}%
}
\caption{Quantitative comparison on class-to-image generation on ImageNet with DiT-XL/2.}
\label{tab:img}
\end{table*}

\begin{figure*}
    \centering
    \includegraphics[width=0.9\linewidth]{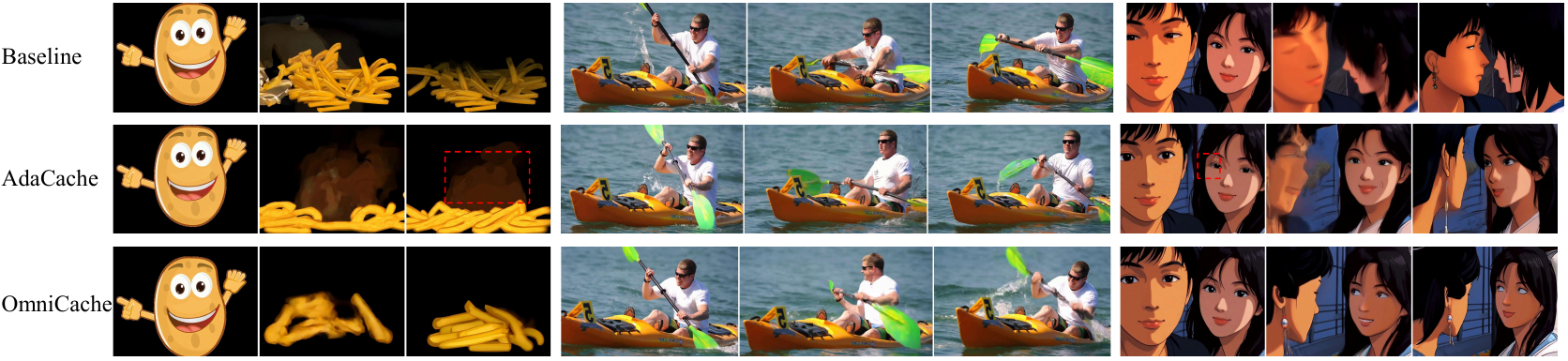}
    \caption{The image-to-video generation results on the CogVideoX-5b-I2V-distill model are shown. The generated video comprises 80 frames, and we extracted the 10th, 40th, and 80th frames for visualization.}
    \label{fig:visual_cogvideo}
\end{figure*}

\subsection{Performance On Video Generation} 

\noindent
\textbf{Text-to-video Generation} Quantitative comparison between our OmniCache with other training-free DiT acceleration methods is shown in Tab.~\ref{tab:t2v}.
In the table, OmniCache-slow and OmniCache-fast perform cache reuse on 50\% and 60\% of the time steps, respectively. For the 30-step Latte model, OmniCache-slow and OmniCache-fast save 15 and 18 steps of computation, respectively. Specifically, our OmniCache-slow achieves a 2$\times$ speedup with only a 0.16\% drop in the VBench metric compared to the baseline, outperforming other training-free, cache-based DiT acceleration methods. Moreover, OmniCache-fast further accelerates the process, striking an excellent balance between performance and acceleration—with a 2.5$\times$ speedup and only a 0.31\% reduction in VBench.

\noindent
\textbf{Image-to-video Generation}
We generated 100 videos using the CogVideoX-5b-I2V-distill model to evaluate the effectiveness of our method. This model is a distilled version with 16 sampling steps, and the resolution of the generated videos is consistent with that of the input images, reaching up to 1080P (768×1360). Due to extensive reduction in redundancy through model distillation, existing cache reuse methods such as $\Delta$-DiT, learn to cache, and PAB fail to generate proper videos, often resulting in model collapse. Shown in Tab.~\ref{tab:cogvideo}, we adapted AdaCache for this model, which performs cache reuse on an average of 4 sampling steps, leading to a 2\% decrease in the Q-Align metric. In contrast, our OmniCache, which accelerates 5 steps (an acceleration ratio of 1.45), is able to maintain a stable qalign metric.
In addition, we conducted human evaluations to compare the baseline and our OmniCache. The win rates for visual quality and textual instruction compliance were \textbf{5\%} and \textbf{-3\%}, respectively. This indicates that our method produces results that are nearly identical to the original model, with almost no degradation in generation quality at an acceleration ratio of 1.45.

We visualized the intermediate frames of the generated videos in Fig.~\ref{fig:visual_cogvideo}. It can be observed that the similarity-based AdaCache introduces additional noise, which cannot be filtered out in subsequent sampling steps, resulting in some blurry regions that degrade the video quality. In contrast, although the final generated videos from our method differ significantly from the original, the overall image quality remains much closer to the original—almost \textbf{lossless}.

\subsection{Performance on Image Generation}

\noindent
\textbf{Class-Conditional Image Generation} Quantitative comparison between OmniCache with other training-free DiT acceleration methods is shown in Tab.~\ref{tab:img}.
DiT-XL/2, based on a DDPM sampler, generates images in 250 steps. Due to the extremely high similarity between adjacent steps, the magnitude of noise induced by cache reuse is lower compared to that in the video generation task, making it challenging for our noise correct module. Nevertheless, our method still achieves results comparable to those of the contemporaneous SOTA method, ToCa.

\begin{table}[t]
\centering
\resizebox{0.47\textwidth}{!}{%
\begin{tabular}{lcc}
\toprule
\textbf{Method} & \textbf{Aesthetic Quality$\uparrow$} & \textbf{Q-Align$\uparrow$} \\ \midrule
CogVideoX-5b-I2V-distill & 0.59  & 0.79  \\
+ OmniCache (Cache Reuse) & 0.58  & 0.778 \\
+ OmniCache (Cache Reuse + Noise Correct) & 0.593 & 0.788 \\
+ OmniCache (Cache Reuse + Noise Correct/Filtering) & 0.621 & 0.792 \\ \bottomrule
\end{tabular}%
}
\caption{The effect of different modules of OmniCache with CogVideoX-5b-I2V-distill.}
\label{tab:ablation_ex}
\end{table}
\subsection{Ablation Study}

Our method, OmniCache uses the sampling trajectory to determine the cache reuse time steps (CacheReuse). And noise correction and filtering module (Noise Correct/Filtering) occurs during sampling stage. The methods listed in Tab.~\ref{tab:ablation_ex} are consistent with those in Tab.~\ref{tab:cogvideo} , all methods perform cache reuse on 5 sampling steps, corresponding to an acceleration ratio of 1.45.
It can be observed that our noise correction module is capable of rectifying cache-induced noise, thereby improving the quality of generated videos. Moreover, when we further apply high-pass and low-pass filtering to the estimated noise to focus on specific information (Filtering), the corresponding video metrics are further enhanced, resulting in nearly lossless performance compared to the original model.
\section{Conclusion and Limitation}


Unlike previous cache reuse methods based on similarity or other local features, our OmniCache provides an novel perspective on diffusion model acceleration through two key innovations. First, we analyze sampling trajectory geometry to determine optimal cache reuse opportunities based on trajectory curvature patterns, enabling informed decisions about when cache reuse will least impact generation quality. Second, we address the noise introduced during cache reuse through dynamic estimation of its correlation with previous sampling steps,  while simultaneously applying high-pass and low-pass filtering at different sampling stages to effectively remove interfering signals while preserving essential components.  Our method proves highly effective in video generation tasks with relatively few sampling steps (30–50 steps), achieving nearly lossless performance with a 2× acceleration. Furthermore, for distilled models with low redundancy where current cache reuse methods almost fail, our approach still delivers video quality that is nearly lossless compared to the original model at an acceleration ratio of 1.45.

OmniCache's main limitation is that cache reuse cannot be applied for three consecutive steps. We introduced this constraint to ensure the reliability of our cache-induced noise estimation, which may limit its maximum acceleration capability.
\subsection{Acknowledgment}
OmniCache was carried out at Zhipu AI. We gratefully acknowledge the CogVideo team and other collaborating groups for their generous support and assistance during research and development. We also thank  the inference infrastructure team for the insightful technical discussions that helped shape this project.

{
    \small
    \bibliographystyle{ieeenat_fullname}
    \bibliography{main}
}

\setcounter{page}{1}
\maketitlesupplementary

\section{Pseudocode of OmniCache}

\begin{algorithm}[htbp]
    \caption{Calibration Stage of OmniCache}
    \label{alg:calib_concise}
    \small
    \begin{algorithmic}[1]
    \REQUIRE  Noisy input $x_T$, model $\epsilon_\theta$
    \ENSURE reuse set $\mathcal{S}$, coeffs $\gamma_t$
    \FOR{$t = T : 1$}
        \STATE $q \gets \widetilde\epsilon_\theta(x_{t},t) -
        \epsilon_\theta(x_{t},t)$
        \STATE Save $q$ and $x_{t}$ 
    \ENDFOR
    \STATE get trajectory/curvature through  saved $x_{t}$ 
    \STATE get coeffs $\gamma_t$ through saved $q$
    \STATE get reuse set $\mathcal{S}$
    \end{algorithmic}
    \label{alg:1}
    \end{algorithm}

\begin{algorithm}[htbp]
    \caption{Inference Stage of OmniCache}
    \label{alg:sampling_concise}
    \small
    \begin{algorithmic}[1]
    \REQUIRE  Noisy input $x_T$, model $\epsilon_\theta$, reuse set $\mathcal{S}$, coeffs $\gamma_t$
    \ENSURE Denoised output $x_0$
    \FOR{$t = T : 1$}
        \IF{$t \in \mathcal{S}$}
            \STATE $q \gets \widetilde\epsilon_\theta(x_{t+1},t+1) -
            \epsilon_\theta(x_{t+1},t+1)$
            \COMMENT{Both are pre-stored on step $t+1$}
            \STATE $\epsilon_\theta(x_t,t)\gets \widetilde\epsilon_\theta(x_t,t)\;-\;\gamma_{t-1}\,\mathrm{Filter}(q,t)$  
            \COMMENT{Filter: LowPass if early, else HighPass}
        \ENDIF
        \STATE Normal Inference Step
    \ENDFOR
    \RETURN $x_0$
    \end{algorithmic}
    \label{alg:2}
    \end{algorithm}

OmniCache operates in two stages: a calibration stage and an inference stage.
\subsection{Calibration Stage}
As shown in Alg.~\ref{alg:1}, we store, for various input examples, the hidden states $x_t$ at different timesteps along with the noise introduced by cache reuse.  Based on these pre‑stored $\{x_t\}$, we reconstruct the diffusion model’s full sampling trajectory and compute its local curvature.  We then select those sampling steps whose omission has the least effect on the trajectory; at exactly these steps we perform cache reuse in order to accelerate inference.  Moreover, since the noise induced by cache reuse at adjacent timesteps tends to be correlated, we use the pre‑stored noise $q$ to estimate an inter‑step noise correlation coefficient $\gamma_t$, which will later guide our noise‑correction procedure.

\subsection{Inference Stage}
During inference, we apply cache reuse at the predetermined “reuse set” $\mathcal{S}$.
 Our cache reuse occurs at each DIT blocks. Regarding the OmniCache inference process, as shown in Alg.~\ref{alg:2}, we perform two forward passes for each non-skipped step t: 1) One actual forward pass, 2) One forward pass with cache reuse at step t. This dual-pass approach allows us to compute the noise $q_t$ introduced by cache reuse at step t. When we intend to skip the subsequent step t-1, we estimate $q_{t-1}$ using $q_t$ and the correlation $\gamma_t$, thereby deriving the corrected output for step t-1.

\section{Latency}
In Tab~\ref{tab:latency}, we report the wall-time breakdown of OpenSora during inference. Out of 30 sampling steps, OmniCache eliminates 15 steps of computation; unlike ToCa and similar methods, we do not rely on any sparsity-based operations. The additional overhead from noise correction and high-/low-pass filtering is negligible compared to a standard sampling step.

\begin{table*}[h!]
\begin{center}
\caption{Latency Comparison in inference stage.}
\resizebox{0.8\linewidth}{!}{%
\begin{tabular}{l c c c c c c}
\toprule
Latency & Original (30 Steps) & \multicolumn{5}{c}{OmniCache‐Slow} \\
\cmidrule(lr){3-7}
    &  & Real Inference & Cache Reuse & High/Low pass  & Noise  & All Inference \\
    &  & Steps (15)     & Steps (15)   & Filtering           & Correction                 & Time            \\
\midrule
OpenSora & 25.72 s & 12.645 s & 0.15 s & $8.4e^{-3}$ s & $3e^{-3}$ s & \textbf{12.80 s} \\
\bottomrule
\end{tabular}%
}
\label{tab:latency}
\end{center}
\end{table*}

\begin{figure*}[b]
    \centering
    \begin{minipage}{0.2\linewidth}
        \centering
        + TeaCache-fast (2.25$\times$)
    \end{minipage}
    \begin{minipage}{0.2\linewidth}
        \centering
        + OmniCache-slow (2$\times$)
    \end{minipage}
    \begin{minipage}{0.2\linewidth}
        \centering
        + OmniCache-fast (2.5$\times$)
    \end{minipage}
    \begin{minipage}{0.2\linewidth}
        \centering
        Original
    \end{minipage}
    \includegraphics[width=0.8\linewidth]{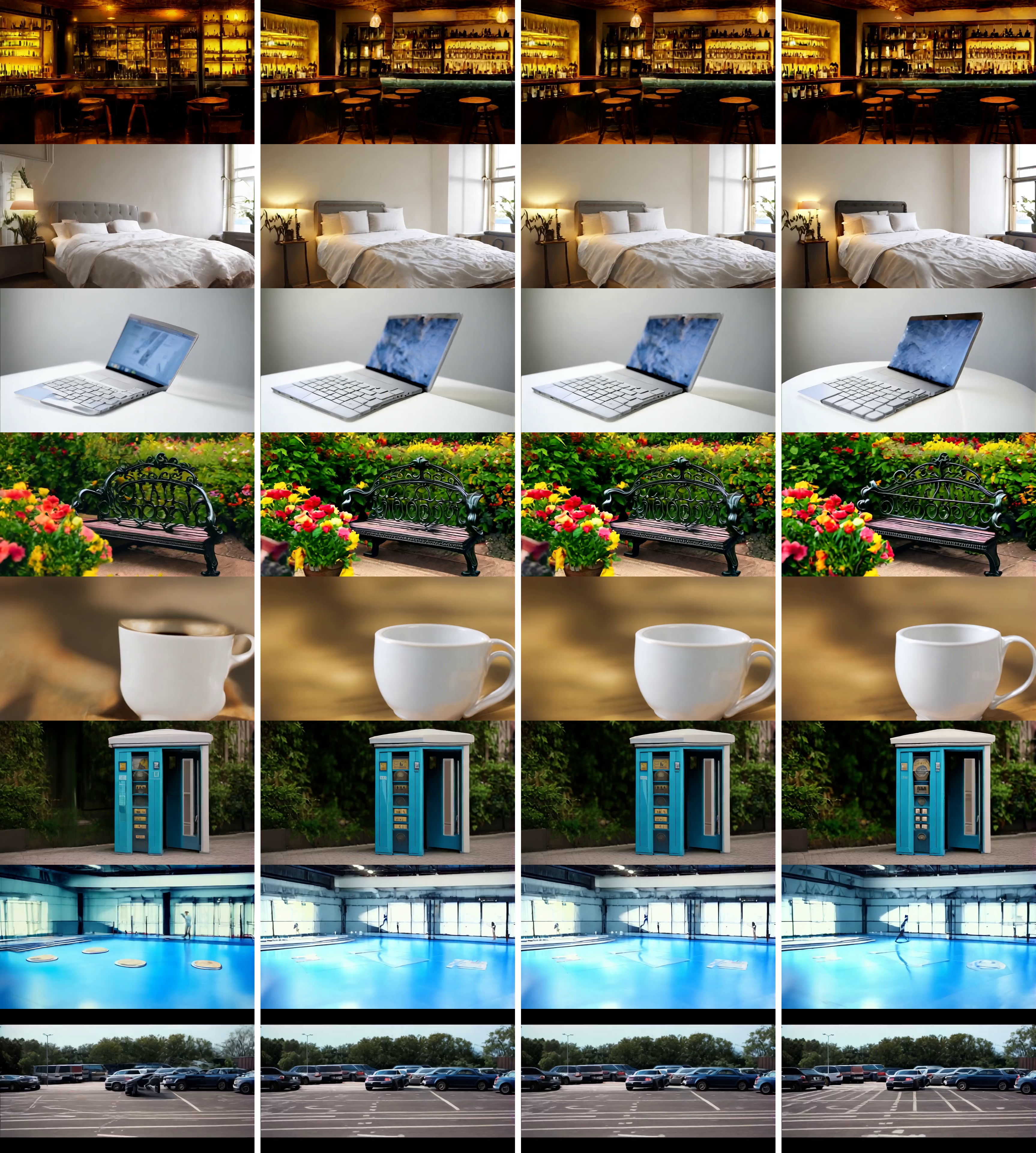}
    \caption{First-Frame Visualization of the Output Video on OpenSora V1.2 (480P, 2s at 30 Steps)}
    \label{fig:opensora}
\end{figure*}

\section{Visualization on OpenSora}
As shown in Fig.~\ref{fig:opensora}, our visual comparison on OpenSora reveals that the current SOTA method, TeaCache, produces outputs with noticeable deviations from the original model, such as the notebook in the third row and the water cup in the fifth row. In contrast, our method generates results that are essentially consistent with the original model while achieving a 2.5× acceleration.

\begin{figure*}[htbp]
    \centering
    \begin{minipage}{0.19\linewidth}
        \centering
        + TeaCache-fast (2.25$\times$)
    \end{minipage}
    \begin{minipage}{0.19\linewidth}
        \centering
        + OmniCache-slow (2$\times$)
    \end{minipage}
    \begin{minipage}{0.19\linewidth}
        \centering
        + OmniCache-fast (2.5$\times$)
    \end{minipage}
    \begin{minipage}{0.19\linewidth}
        \centering
        Original
    \end{minipage}
    \includegraphics[width=0.76\linewidth]{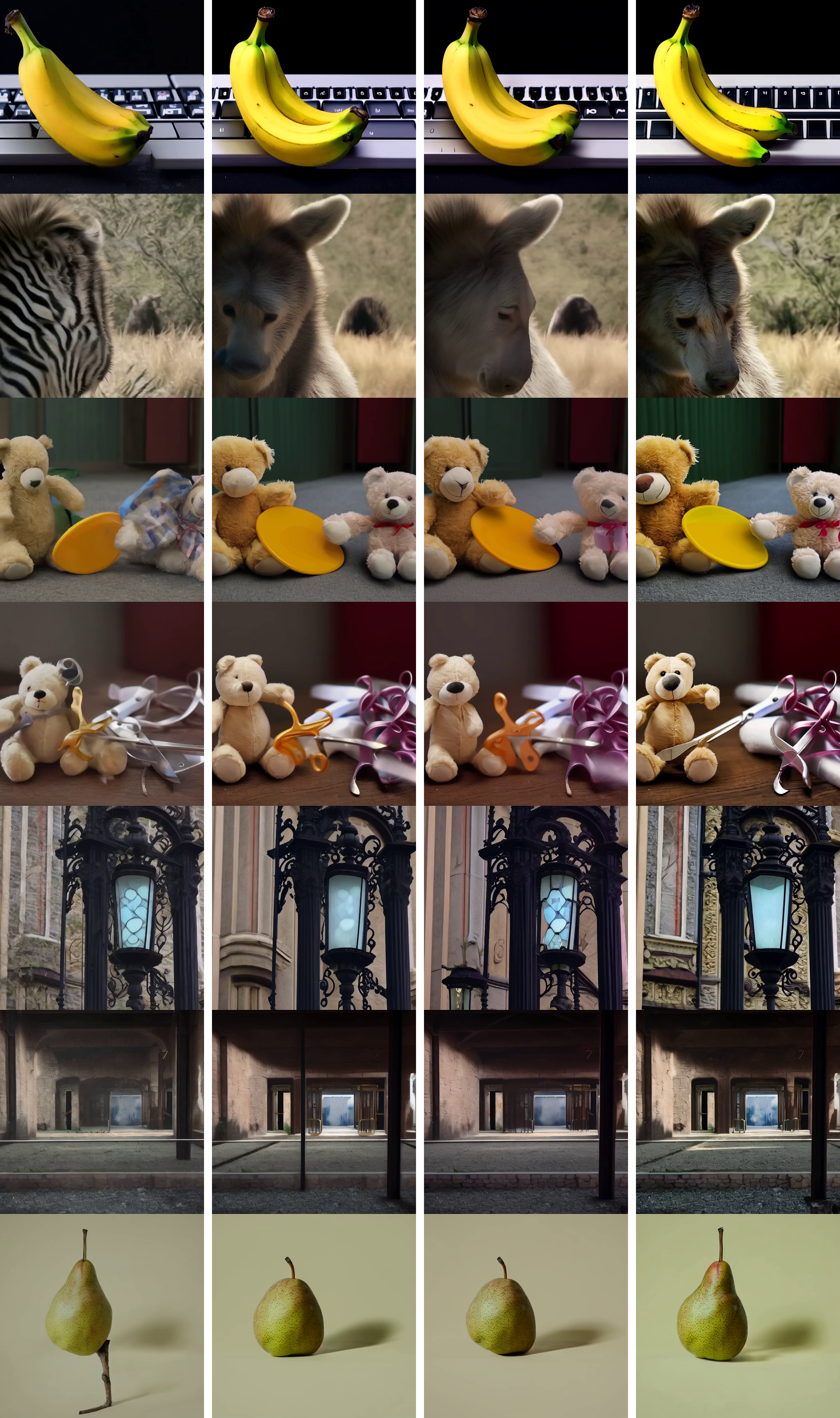}
    \caption{First-Frame Visualization of the Output Video on Latte (512$\times$512, 2s at 50 Steps)}
    \label{fig:latte}
\end{figure*}
\section{Visualization on Latte}
As shown in Fig.~\ref{fig:latte}, we conducted a visual comparison on Latte. TeaCache exhibits many abnormal distortions—for example, the pear in the last row and the teddy bear in the fifth row. In contrast, our generated videos are more consistent with the original model, and the objects in the videos appear more natural.



\end{document}